\documentclass[conference]{IEEEtran}
\IEEEoverridecommandlockouts
\usepackage{cite}
\usepackage{amsmath,amssymb,amsfonts}
\usepackage{algorithm,algorithmicx}
\usepackage{algpseudocode}

\newtheorem{Thm}{Theorem}[section]

\newtheorem{claim}[Thm]{Claim}

\usepackage{graphicx}
\usepackage{xcolor}
\usepackage{tikz}
\usetikzlibrary{calc}
\usepackage{comment}
\usepackage{bm}
\def\BibTeX{{\rm B\kern-.05em{\sc i\kern-.025em b}\kern-.08em
    T\kern-.1667em\lower.7ex\hbox{E}\kern-.125emX}}

\algnewcommand{\Initialize}[1]{%
  \State \textbf{Initialize:}
  \State \hspace*{\algorithmicindent}\parbox[t]{0.8\linewidth}{\raggedright #1}
}

\begin{document}

\title{Stabilizing Private LASSO under Heterogeneous Covariates via Anisotropic Objective Perturbation
\thanks{This work is partially supported by JSPS KAKENHI
(22H05117) and JST PRESTO (JPMJPR23J4).}
}

\author{\IEEEauthorblockN{Haruka Tanzawa}
\IEEEauthorblockA{\textit{Department of Information Science} \\
\textit{Ochanomizu University}\\
Bunkyo-ku, Tokyo, Japan \\
g2120528@is.ocha.ac.jp}
\and
\IEEEauthorblockN{Ayaka Sakata}
\IEEEauthorblockA{\textit{Department of Information Science} \\
\textit{Ochanomizu University}\\
Bunkyo-ku, Tokyo, Japan }
\IEEEauthorblockA{\textit{RIKEN center for AIP} \\
Chuo-ku, Tokyo, Japan \\
ayakasakata@is.ocha.ac.jp}
}

\maketitle

\begin{abstract}

We study high-dimensional LASSO under differential privacy via objective perturbation with heterogeneous covariate scales.
In practical scenarios, covariates often exhibit diverse scales; however, standard preprocessing is problematic under privacy constraints, as it consumes additional privacy budget.
This heterogeneity induces effective anisotropy in the objective perturbation via the inverse Gram matrix of covariates, which can degrade the stability and accuracy of algorithms.
To address this, we propose a Gram-based anisotropic objective perturbation, a ``pre-distortion" strategy that counteracts the distortion from the covariate structure to restore isotropy in the estimation process. Using an Approximate Message Passing (AMP) framework and state evolution analysis, we demonstrate that our proposed perturbation 
significantly stabilizes convergence and improves both statistical efficiency and privacy performance compared to standard uniform noise injection.
Our results provide theoretical insights into designing stable and efficient private estimators without relying on data-dependent preprocessing.

\end{abstract}


\section{Introduction}

Recent concerns over data privacy have established Differential Privacy (DP) as the gold standard for protecting sensitive information \cite{dwork2006calibrating}.
A common approach to achieve differential privacy is to randomize the process of parameter estimation via a \textit{randomized mechanism}, where the injected noise conceals the contribution of individual data points used in training.
Several such mechanisms have been proposed, including output perturbation, where noise is added directly to the estimated parameters \cite{dwork2006calibrating}, and objective perturbation, where noise is injected into the loss function before optimization \cite{chaudhuri2011differentially}.
A fundamental challenge in all such approaches is balancing the privacy-utility trade-off \cite{dwork2014algorithmic}: stronger privacy guarantees typically require more randomness, which can deteriorate statistical accuracy.

In modern applications such as genomics and electronic health records, where both high-dimensionality and privacy protection are critical \cite{naveed2015privacy}, sparse regression techniques play a central role \cite{tibshirani1997lasso}.
In such settings, understanding the interplay between sparsity and privacy is an important theoretical problem \cite{talwar2015nearly}.
Recent work has employed Approximate Message Passing (AMP) to analyze the asymptotic behavior of regularized estimators under randomized mechanisms \cite{sakata2026privacy}.
This study has shown that randomized mechanism defined by objective perturbation, while preserving sparsity, can induce instability in AMP, degrading both estimation and privacy performance when the noise level becomes excessively large.

In many practical settings, covariates exhibit heterogeneous scales and correlations, deviating from idealized i.i.d.\ assumptions. While such heterogeneity is typically mitigated by preprocessing (e.g., standardization or whitening) in non-private settings, these operations are problematic under differential privacy as they rely on global data statistics and incur additional privacy costs \cite{dwork2014algorithmic}.
As a result, one must handle heterogeneous covariates directly, making noise design particularly important. Prior work has shown that anisotropic, covariance-aware noise can significantly improve utility by aligning with the geometry of the problem \cite{nikolov2013geometry}, in the context of output perturbation, suggesting that such designs are essential when preprocessing is not feasible.

However, such settings remain poorly understood in high-dimensional asymptotic analysis under objective perturbation.
In particular, AMP-based frameworks typically assume homogeneous covariates, and do not capture the effects of heterogeneous covariate and the corresponding anisotropic perturbations.
Given that instability can arise even under isotropic noise \cite{sakata2026privacy}, the presence of heterogeneity is expected to further affect convergence behavior and estimation performance, but its precise impact remains largely unexplored.

Motivated by these observations, we study anisotropic perturbations in high-dimensional LASSO under objective perturbation with heterogeneous covariates.
We develop an AMP-based framework to analyze the effect of anisotropic perturbations on convergence and estimation performance, and propose a component-wise noise allocation strategy that adapts to heterogeneous covariate scales.
Our analysis reveals how noise anisotropy affects estimation accuracy, convergence behavior, and privacy guarantees, and shows that the proposed approach improves both stability and statistical efficiency compared to uniform noise injection.

\subsection{Our contribution}

Our contributions are summarized as follows:
\begin{itemize}

    \item We show that objective perturbation inherently induces anisotropic noise: 
    even if the injected noise is isotropic, the inverse Gram matrix of the covariates amplifies perturbations along high-sensitivity directions. 
    This implicit anisotropic perturbation leads to instability of parameter estimation. 

    \item To this end, we propose a Gram-based anisotropic objective perturbation that compensates for the distortion induced by the inverse Gram matrix, partially suppressing excessive noise amplification in sensitive directions.

    \item We develop an AMP algorithm with component-wise noise allocation under heterogeneous covariates, and show that the proposed anisotropic perturbation significantly stabilizes convergence and improves generalization performance compared to uniform noise.

     \item We derive state evolution equations that characterize the asymptotic dynamics of AMP under heterogeneous covariate scales and noise levels, and demonstrate that anisotropic noise improves the privacy-utility trade-off.
\end{itemize}

\section{Randomization with Objective perturbation}

Differential privacy requires that the output distribution be insensitive to a change in a single data point. 
Let ${\cal D}_\mu^\prime$ denote a one-point-mutant (OPM) dataset that differs from ${\cal D}$ only in the $\mu$-th sample. 
A common approach to achieve this is to introduce randomness via a randomized mechanism so that the output distributions under ${\cal D}$ and ${\cal D}_\mu^\prime$ are close.

One such mechanism is given by objective perturbation, which we adopt in this paper. 
Specifically, we consider the LASSO with objective perturbation \cite{chaudhuri2011differentially,sakata2026privacy}:
\begin{align}
    \widehat{\bm{x}}({\cal D},\bm{\eta})=\mathop{\mathrm{argmin}}_{\bm{x}}\biggl\{\frac{1}{2}\|\bm{y}-F\bm{x}\|_2^2+\lambda\|\bm{x}\|_1+\bm{\eta}^\top\bm{x}\biggr\},
    \label{eq:PrivateLASSO}
\end{align}
where ${\cal D}=\{\bm{y},F\}$, $\lambda>0$ is the regularization parameter that controls the
sparsity of the estimate, and $\bm{\eta}$ denotes the objective perturbation, a privacy-preserving noise.

\subsection{Privacy criterion and the privacy--accuracy trade-off}
\label{sec:criterion}

Within this formulation, the privacy requirement can be restated as requiring the
distributions of $\widehat{\bm{x}}({\cal D},\bm{\eta})$ and
$\widehat{\bm{x}}({\cal D}^\prime_\mu,\bm{\eta})$, induced by the randomness of
$\bm{\eta}$, to be close to each other.
To quantify this distributional closeness, we adopt the
On-Average KL (OnAveKL) divergence \cite{wang2016average}, which measures the
distinguishability of individual samples and has been used to characterize vulnerability
to membership inference attacks \cite{shokri2017membership}.
Let $\mathbb{P}(\bm{x}|{\cal D})$ denote the distribution of
$\widehat{\bm{x}}({\cal D},\bm{\eta})$ induced by $\bm{\eta}$. 
The OnAveKL divergence is defined as
\begin{align}
\mathrm{OnAveKL}
=
\frac{1}{M}\sum_{\mu=1}^M
\mathbb{E}_{{\cal D},{\cal D}_\mu^\prime}
\left[
\mathrm{KL}\bigl(
p(\bm{x}|{\cal D})
\|
p(\bm{x}|{\cal D}_\mu^\prime)
\bigr)
\right].
\label{eq:def_OnAveKL}
\end{align}
A smaller OnAveKL value indicates that the output distribution is less
sensitive to changes in individual data points, and hence stronger
resistance to membership inference attacks. Increasing the noise level
is generally expected to reduce OnAveKL and
improve privacy, but may also increase the generalization error. This
potential tension between privacy and accuracy motivates the
privacy--accuracy trade-off considered in this paper.

\subsection{Varying covariate strength and objective perturbation}

To illustrate the effect of varying covariate strength, 
we note that objective perturbation can be interpreted as shifting the center of the quadratic loss, here we denote $\bm{\mu}$. 
In particular, the perturbation enters the estimator through the linear transformation $(F^\top F)^{-1}$.
Assuming isotropic noise $\bm{\eta} \sim \mathcal{N}(\bm{0}, \sigma^2_\eta I)$, this induces 
$\bm{\mu}
\sim
\mathcal{N}\big(
(F^\top F)^{-1}F^\top\bm{y},
\,\sigma_\eta^2 (F^\top F)^{-2}
\big)$.
Consequently, even when isotropic noise is used for privacy, the estimator behaves as if it were subject to anisotropic perturbations determined by the inverse Hessian. 
In particular, directions corresponding to small eigenvalues of $F^\top F$ are subject to amplified noise, as shown in top panels of Fig.\ref{fig:noise_dist}.

\subsection{Effective Anisotropy in Objective Perturbation}
\label{sec:Gram-based_Noise}

\begin{figure}

\begin{center}
\begin{tikzpicture}

\def\NoiseLabel{2.5}
\def\w{2.}
\def\R{1.}
\def\gap{0.5}
\def\vgap{0.5}

\draw (0,0) rectangle (\w,\w);

\draw (\w+\gap,0) rectangle (2*\w+\gap,\w);

\draw (0,-\w-\vgap) rectangle (\w,-\vgap);

\draw (\w+\gap,-\w-\vgap) rectangle (2*\w+\gap,-\vgap);


\draw[blue,thick] (\R,\R) circle (0.8);

\begin{scope}[shift={( \w+\gap+\R , \R)}, rotate=45]
    \draw[red,thick] (0,0) ellipse (1 and 0.5);
\end{scope}

\begin{scope}[shift={(\R , -\gap-\R)}, rotate=-45]
    \draw[red,thick] (0,0) ellipse (1 and 0.5);
\end{scope}

\draw[blue,thick] (\w+\gap+\R , -\gap-\R) circle (0.8);

\draw[-]
    (-2.2,-0.25) -- (5.,-0.25);
\draw[-]
    (-2.2,\w+0.2) -- (5.,\w+0.2);
\draw[-]
    (-2.2,-\w-\vgap-0.2) -- (5.,-\w-\vgap-0.2);
\draw[-]
    (-0.2,-\w-\vgap-0.2) -- (-0.2,\w+\vgap+0.3);
\draw[-]
    (\w+0.5*\gap,-\w-\vgap-0.2) -- (\w+0.5*\gap,\w+\vgap+0.3);


\node at (1.,\NoiseLabel) {Original};
\node at (\w+\gap+1,\NoiseLabel) {Effective};
\node at (-1.2, 1.3) {Isotropic};
\node at (-1.2,0.9) {Perturbation};
\node at (-1.2,-1.3) {Gram-based};
\node at (-1.2,-1.7) {Perturbation};

\end{tikzpicture}
\end{center}
\caption{Schematic of noise distributions under objective perturbation.
Left and right columns display the initially injected noise and the resulting effective noise after transformation by the inverse Gram matrix, respectively.
(Top) Isotropic perturbation, which becomes effectively anisotropic.
(Bottom) Gram-based perturbation, where the injected noise is pre-anisotropic to ensure the effective noise is approximately isotropic.}
\label{fig:noise_dist}
\end{figure}

The induced anisotropy can severely degrade the stability of algorithms such as AMP, whose convergence is known to depend sensitively on the spectral properties of the design matrix \cite{caltagirone2014convergence, rangan2019convergence}. 
While prior work \cite{sakata2026privacy} has shown that even isotropic noise can hinder convergence, the additional anisotropy introduced through the inverse Gram matrix remains largely unexplored, especially under heterogeneous covariate scales where noise amplification becomes highly directional.

To address this issue, we propose Gram-based anisotropic objective perturbation, in which the injected noise is pre-distorted according to the covariate structure so that the resulting effective perturbation becomes approximately isotropic (bottom panels of Fig.~\ref{fig:noise_dist}). 
We analyze the AMP behavior under standard isotropic noise to characterize instability, and then demonstrate that the proposed design significantly improves both convergence and estimation accuracy.

\section{AMP under heterogeneous covariate and effect of privacy noise}

We study AMP for solving \eqref{eq:PrivateLASSO} under heterogeneous covariates. 
While standard AMP assumes i.i.d.\ Gaussian designs \cite{donoho2009message,zdeborova2016statistical,sakata2023prediction}, our goal is to characterize the interplay between covariate scale heterogeneity and privacy noise. 
To retain analytical tractability while capturing the key effect of Gram-based perturbation, we consider a model with independent but heterogeneously scaled covariates.

\subsection{Data Model and Privacy Noise Allocation}

We model heterogeneity as $F_{\mu i}\! =\! \tilde{F}_{\mu i} {s_v}_i$ for $\mu\!\in\!\{1,\ldots,M\}$ and $i\in\{1,\ldots,N\}$, where $\tilde{F}_{\mu i} \sim \mathcal{N}(0, 1/N)$ and ${s_v}_i$ represents the scale of the $i$-th covariate. 
The scales $\{{s_v}_i\}_{i=1}^N$ are i.i.d.\ samples from a distribution $P_v$ with bounded second moments. 
The observations follow $\bm{y} = F\bm{x}^{(0)} + \bm{\xi}$, where $\bm{x}^{(0)}\!\in\mathbb{R}^N$ is a sparse signal whose components are independently drawn from a Bernoulli--Gaussian prior,
$\phi_x(x^{(0)})=(1-\rho)\delta(x^{(0)})+\rho{\cal N}(x^{(0)};0,1)$,
and $\bm{\xi} \sim \mathcal{N}(0, \sigma_\xi^2 I_M)$ is the observation noise.
Here, we consider the high-dimensional regime where $N,M\gg 1$ and $\alpha = M/N\sim O(1)$.

In this setting, $(F^\top F)_{ii} \approx \alpha v_i$ with $v_i={s_v}_i^2$, making the inverse-Gram distortion in
Sec.~\ref{sec:Gram-based_Noise} coordinate-wise. 
Motivated by this, we introduce the Gram-based perturbation, which leads to a component-wise noise allocation:
\begin{align}
    \sigma_{\eta,i}^2=\frac{v_i^2}{\overline{v^2}}\,\overline{\sigma_\eta^2},\quad \overline{v^2}=\frac{1}{N}\sum_{j=1}^Nv_j^2
\label{eq:def_GramBasedNoise}
\end{align}
where $\overline{\sigma_\eta^2}$ is the target average noise level, ensuring $\sum_{i=1}^N\sigma_{\eta,i}^2 = N\overline{\sigma_\eta^2}$. This choice provides an AMP-tractable realization of the proposed pre-distortion strategy.

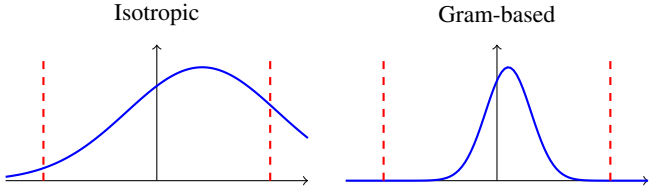
\begin{figure}[t]
\begin{tikzpicture}

\def\xmax{2}
\def\th{1.5}

\begin{scope}
    \draw[->] (-\xmax,0) -- (\xmax,0);
    \draw[->] (0,0) -- (0,1.8);

    \draw[blue, thick, domain=-2:2, samples=100]
        plot (\x,{1.5*exp(-0.5*(\x/1.-0.6)^2)});

    \draw[red, thick, dashed] (\th,0) -- (\th,1.6);
    \draw[red, thick, dashed] (-\th,0) -- (-\th,1.6);

    \node at (0,2.2) {\small Isotropic};
\end{scope}

\begin{scope}[xshift=4.5cm]
    \draw[->] (-\xmax,0) -- (\xmax,0);
    \draw[->] (0,0) -- (0,1.8);

    \draw[blue, thick, domain=-2:2, samples=100]
        plot (\x,{1.5*exp(-0.5*((\x-0.15)/0.3)^2)});

    \draw[red, thick, dashed] (\th,0) -- (\th,1.6);
    \draw[red, thick, dashed] (-\th,0) -- (-\th,1.6);

    \node at (0,2.2) {\small Gram-based};
\end{scope}

\end{tikzpicture}
\caption{Relationship between the threshold (dashed vertical lines) and the distribution of the pre-thresholding estimate.
Under isotropic perturbation, components with small $v$ (i.e., high sensitivity) exhibit a large variance due to amplified noise, resulting in significant overlap with the threshold.
In contrast, under Gram-based perturbation, the distribution is more concentrated while the threshold remains large, leading to more reliable shrinkage to zero.
}
\label{fig:threshold_vs_perturbation}
\end{figure}

\subsection{AMP Algorithm and Effective Thresholding under Gram-based perturbation}

Algorithm~\ref{alg:AMP} presents AMP under heterogeneous covariate scales, 
where $\circ$ and $\oslash$ denote component-wise product and division, respectively, and $\bm{1}_N$ is the $N$-dimensional all-ones vector. 
Objective perturbation $\bm{\eta}$ is incorporated through the scalar denoising step \cite{sakata2026privacy}, with
\begin{align}
    \mathbb{M}(\Sigma,\mathrm{m})
    &=\left(\mathrm{m}-\mathrm{sgn}(\mathrm{m})\lambda\Sigma\right)\mathbb{I}(|\mathrm{m}|>\lambda\Sigma), \label{eq:AMP_M}\\
    \mathbb{V}(\Sigma,\mathrm{m})
    &=\Sigma\,\mathbb{I}(|\mathrm{m}|>\lambda\Sigma),
\end{align}
where $\mathbb{I}(\cdot)$ is the indicator function.
In the decoupled scalar channel, the effective variance scales as $\Sigma \propto v^{-1}$. 
Thus, components with smaller $v$ are associated with larger thresholds $\lambda \Sigma$. 
The pre-thresholding estimate takes the form $\mathrm{m} - \eta \Sigma$ (Algorithm~\ref{alg:AMP}). 
If $\eta$ is independent of $v$, both the noise term $\eta \Sigma$ and the threshold scale as $v^{-1}$, amplifying fluctuations in low-$v$ directions
(Fig.~\ref{fig:threshold_vs_perturbation}, left).
In contrast, the proposed scaling $\eta \propto v$ makes $\eta \Sigma=O(v^0)$, while the threshold $\lambda \Sigma$ remains large for small $v$. 
As a result, components with weak covariate scales are more likely to be shrunk to zero (Fig.\ref{fig:threshold_vs_perturbation} right).

This property is also desirable from a privacy perspective. 
Since sensitivity, defined as the change in the estimator induced by the replacement of a single data point, is governed by the inverse Gram matrix, directions with small $v$ are more vulnerable to leakage. 
The Gram-based perturbation suppresses these high-sensitivity components, effectively ``hiding'' them behind large thresholds, thereby mitigating privacy leakage.

\begin{algorithm}[t]
\caption{AMP for Heterogeneous Covariates}
\label{alg:AMP}
\begin{algorithmic}[1]
\Require {${\cal D}=\{\bm{y},F\}$}
\Ensure {$\widehat{\bm{x}}({\cal D},\bm{\eta})$}
    \State{$\bm{v}\gets\alpha^{-1}\mathrm{diag}(F^\top F)$}\Comment{Column-wise scale}

    \State{$\widehat{\bm{x}}^{(1)}\gets$ draw initial values from ${\cal N}(0,1)$}
    \State{$\bm{s}^{(1)}\gets$ draw initial values from $[0,1]^N$}
    \State{$\bm{g}_{\mathrm{out}}^{(0)}\gets\bm{0}_M$}
    \For{$t=1$ to $T$}
    \State{$\widehat{\bm{\theta}}^{(t)}\gets\bm{F}\widehat{\bm{x}}^{(t)}-\bm{g}_{\mathrm{out}}^{(t-1)}\circ\left((\bm{F}\circ\bm{F})\bm{s}^{(t)}\right)$}
\State{$s_\theta^{(t)}\gets\bm{v}^\top\bm{s}^{(t)}\slash N$}
\State{$\bm{g}_{\mathrm{out}}^{(t)}\gets \dfrac{\bm{y}-\widehat{\bm{\theta}}^{(t)}}{1+s_\theta^{(t)}}$}
\State{$\bm{\Sigma}^{(t)}\gets 
\alpha^{-1}\left({1+s_\theta^{(t)}}\right)\bm{1}_N\oslash\bm{v}$}
\State{$\bm{\mathrm{m}}^{(t)}\gets\bm{\Sigma}^{(t)}\circ
\left(\bm{F}^{\top}\bm{g}_{\mathrm{out}}^{(t)}+
\dfrac{\alpha\widehat{\bm{x}}^{(t)}\circ\bm{v}}{1+s_\theta^{(t)}}\right)$}
\State{$\widehat{\bm{x}}^{(t+1)}\gets \mathbb{M}(\bm{\Sigma}^{(t)},\bm{\mathrm{m}}^{(t)}-\bm{\eta}\circ\bm{\Sigma}^{(t)})$}
\State{$\bm{s}^{(t+1)}\gets \mathbb{V}(\bm{\Sigma}^{(t)},\bm{\mathrm{m}}^{(t)}-\bm{\eta}\circ\bm{\Sigma}^{(t)})$}
    \EndFor
    \State \Return $\widehat{\bm{x}}^{(T+1)}$

\end{algorithmic}
\end{algorithm}

\subsection{Results of AMP: Generalization and Convergence}

We compare the behavior of AMP under Gram-based and isotropic perturbations. 
In the Gram-based perturbation, we set $\eta_i\sim{\cal N}(0,\,\sigma_{\eta,i}^2)$.
To model heterogeneity, we consider the following distributions for the covariate scale $s_v$:
\begin{itemize}
    \item[\textbf{v1}:] Uniform distribution on $(0,1]$
    \item[\textbf{v2}:] Log-normal distribution with log-mean $0$ and log-variance $0.5^2$
\end{itemize}
Under this formulation, the noise strength \eqref{eq:def_GramBasedNoise} can be written as a function of $v$: $\sigma_\eta^2(v)=v^2\,\overline{\sigma_\eta^2}\slash \mathbb{E}_v[v^2]$.

\begin{figure}
\centering
    \includegraphics[width=\columnwidth]{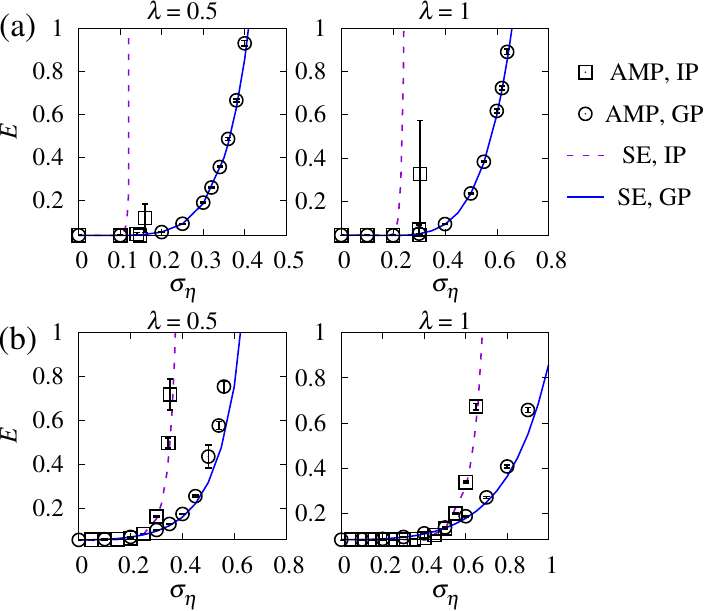}
    \caption{
    Generalization error vs.\ privacy noise strength for (a) uniform $v$ (\textbf{v1}) and (b) log-normal $v$ (\textbf{v2}) at $\alpha=0.5$, $\rho=0.1$ and $\sigma_\xi=0.1$. AMP at $N=1000$ and state evolution (SE) results are shown for isotropic (IP) and Gram-based (GP) perturbations.
    }
    \label{fig:AMP_SE_alpha0d5_rho0d1}
\end{figure}

Fig.~\ref{fig:AMP_SE_alpha0d5_rho0d1} shows the generalization error $E$ at $\alpha=0.5$ and $\rho=0.1$ as a function of the noise strength $\sigma_\eta := \sqrt{\overline{\sigma_\eta^2}}$ for (a) \textbf{v1} and (b) \textbf{v2}. 
Under isotropic perturbation, AMP fails to converge beyond a certain noise level, resulting in missing data points. 
In contrast, the Gram-based perturbation remains stable even at higher noise levels, and the increase of $E$ with respect to noise is more gradual. 
These results demonstrate that the Gram-based perturbation improves both the convergence of AMP and generalization under heterogeneous covariates. Similar trends are observed for other values of $\alpha$ and $\rho$.

In Fig.~\ref{fig:E_and_rho_vs_lambda}(a), we show $E$ versus $\lambda$
at $\alpha=0.5$, $\rho=0.1$, and $\sigma_\eta=0.1$. The Gram-based
perturbation permits smaller $\lambda$ than the isotropic one, with the
difference most pronounced in the uniform setting (\textbf{v1}), where
frequent low-$v$ components induce instability that is mitigated by the
proposed method. For large $\lambda$, however, the impact of perturbation
design diminishes as sparsity itself stabilizes the estimator.

\section{State Evolution Analysis}

To gain further theoretical insight into the behavior of AMP, 
we introduce the state evolution (SE) analysis, 
which characterizes the typical behavior of the AMP trajectories \cite{zdeborova2016statistical}.

\subsection{Decoupling principle under varying covariate}

We show that the decoupling principle \cite{bayati2011dynamics} extends to the heterogeneous covariate scales $\bm{v}$:
each component depends
explicitly only on its corresponding $(x_i^{(0)},v_i,\eta_i)$, while
the remaining components affect it only through macroscopic
quantities.
More precisely, for a given covariate scale $v$, ground truth $x^{(0)}$, and objective perturbation $\eta$,
we denote the resulting scalar estimator by 
$\widehat{x}^{(t)}(\tilde{F} \mid x^{(0)}, v,\eta)$.
\begin{claim}[Decoupling principle with heterogeneous scales]
Assume that the covariate scales $\{v_i\}$ are i.i.d.\ positive random variables 
with finite second moments. 
Then, in the high-dimensional limit $M,N \to \infty$ with $\alpha = M/N = O(1)$, 
the AMP estimator under privacy noise satisfies
\begin{align}
    \widehat{x}^{(t+1)}(F|x^{(0)},v,\eta)\overset{d}{=}\mathbb{M}\left(\Sigma^{(t)}_v,x^{(0)}+\sigma_z^{(t)}z-\eta\Sigma_v^{(t)}\right),
    \label{eq:AMP_equivalence}
\end{align}
where $\sigma_{z}^{(t)}=\sqrt{E^{(t)}/(\alpha v)}$, 
$\Sigma_v^{(t)}=(1+V^{(t)})/(\alpha v)$, and $z\sim{\cal N}(0,1)$, and the parameters evolve according to
\begin{align*}
    E^{(t+1)}\!&=\left\langle \! v\!\left(x^{(0)}\!-\!\mathbb{M}\left(\Sigma_v^{(t)},x^{(0)}+\sigma_zz-\!\eta\Sigma_v^{(t)}\right)\right)^2\right\rangle\!+\!\sigma_\xi^2\\
    V^{(t+1)}\!&=\left\langle v\mathbb{V}\!\left(\Sigma_v^{(t)},x^{(0)}+\sigma_zz-\eta\Sigma_v^{(t)}\right)\!\right\rangle
\end{align*}
and 
$
    \left\langle a(\cdot)\right\rangle=\mathbb{E}_{v,x^{(0)},z}\left[\int d\eta ~{\cal N}(\eta;0,\sigma_{\eta}^2(v))a(\cdot)\right].
$

\end{claim}
The heuristic derivation is provided in Appendix~\ref{app:state_evolution}.

In Fig.~\ref{fig:AMP_SE_alpha0d5_rho0d1}, we compare SE with AMP for
both isotropic and Gram-based perturbations. The close agreement
supports our theoretical analysis.

\begin{figure}[t]
    \centering
    \includegraphics[width=\columnwidth]{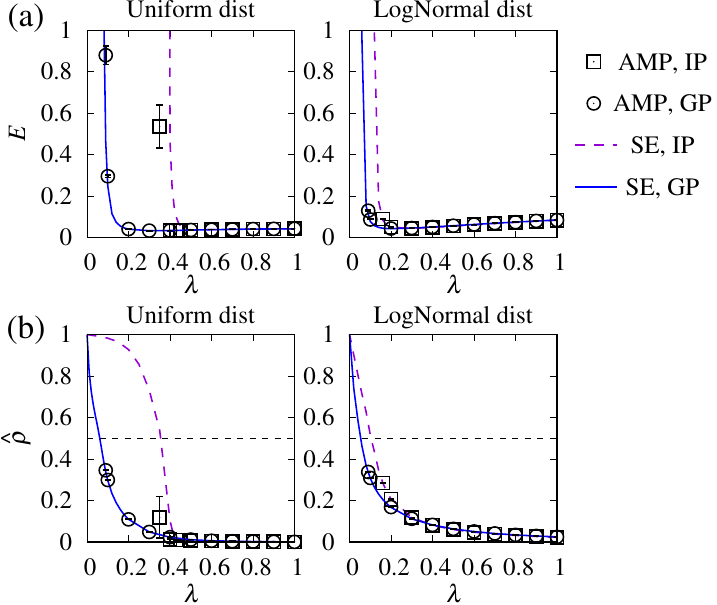}
    \caption{$\lambda$-dependence of (a) the generalization error $E$ and (b) the fraction of non-zero components $\widehat{\rho}$ at $\alpha=0.5,\rho=0.1, \sigma_\xi=0.1$, and $\sigma_\eta=0.1$ for \textbf{v1} (Uniform) and \textbf{v2} (LogNormal). The legend is the same as in Fig.~3. The horizontal lines in (b) shows $\widehat{\rho}=0.5$}
    \label{fig:E_and_rho_vs_lambda}
\end{figure}

Using SE, we derive a stability condition for AMP: it typically
converges when $\mathbb{E}_{{\cal D},\bm{\eta}}[\widehat{\rho}({\cal D},\bm{\eta})]<\! \alpha$ \cite{kabashima2009typical,sakata2023prediction} (see Appendix~\ref{sec:app_AT}), where $\widehat{\rho}$ denotes the fraction of non-zero components in the estimate.
Fig.~\ref{fig:E_and_rho_vs_lambda}(b)
shows $\mathbb{E}_{{\cal D},\bm{\eta}}[\widehat{\rho}({\cal D},\bm{\eta})]$ by AMP versus $\lambda$, together with a horizontal line at $\alpha(=0.5)$. 
AMP fails to converge once $\widehat{\rho}$ exceeds this threshold, as indicated by the absence of data points in the plot.

\subsection{Privacy Analysis via On-Average KL Divergence}

In the high-dimensional limit, AMP decomposes OnAveKL
\eqref{eq:def_OnAveKL} into marginal KL divergences, termed cwOnAveKL,
which can be evaluated using SE.
Focusing on the component-wise marginals $p_i(x_i|{\cal D})$, we analyze:
\[
\mathrm{cwOnAveKL} = \frac{1}{M}\sum_{\mu=1}^M \sum_{i=1}^N \mathbb{E}_{{\cal D},{\cal D}_\mu^\prime}\!\! \left[ {\mathrm{KL}} \big( p_i(x_i|{\cal D}) \big| p_i(x_i|{\cal D}_\mu^\prime) \big) \right].
\]
Utilizing the AMP cavity messages as an approximation of leave-one-out estimators \cite{mezard2009information}, we derive the asymptotic cwOnAveKL via SE for heterogeneous covariate scales \cite{sakata2026privacy}:
\begin{align}
    \mathrm{cwOnAveKL}=\alpha^{-2}E~\mathbb{E}_{v,z,x^{(0)}}\left[R_v\slash v\right]
\end{align}
where $\widehat{\mathrm{m}}_v=x^{(0)}+\sqrt{E\slash (\alpha v)}z$ and
\begin{align*}
    R_v&=\frac{1}{1-\widehat{r}_v}\left(\frac{\partial\widehat{r}_v}{\partial\widehat{\mathrm{m}}_v}\right)^2+\frac{\partial^2\widehat{r}_v}{\partial\widehat{\mathrm{m}}_v^2}+\frac{\widehat{r}_v}{\Sigma_v^2\sigma_\eta^2(v)},\\
    \widehat{r}_v&=\frac{1}{2}\left\{\mathrm{erfc}\left(\frac{-\widehat{\mathrm{m}}_v+\lambda\Sigma_v}{\sqrt{2}\Sigma_v\sigma_\eta(v)}\right)+\mathrm{erfc}\left(\frac{\widehat{\mathrm{m}}_v+\lambda\Sigma_v}{\sqrt{2}\Sigma_v\sigma_\eta(v)}\right)\right\}.
\end{align*}

\begin{figure}
    \centering
    \includegraphics[width=3in]{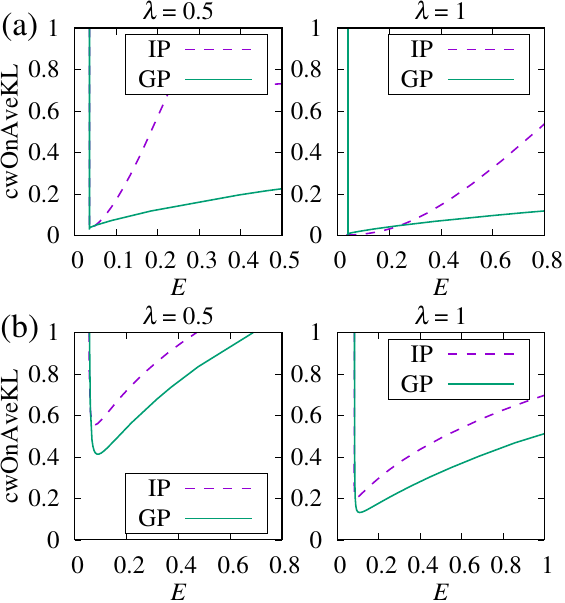}
    \caption{Privacy-accuracy trade-off at $\alpha=0.5$, $\rho=0.1$, and $\sigma_\xi=0.1$ for (a) setting \textbf{v1} and (b) \textbf{v2}. IP denotes isotropic perturbation and GP denotes Gram-based perturbation.}
    \label{fig:E_vs_cwOnAveKL}
\end{figure}

Fig.~\ref{fig:E_vs_cwOnAveKL} shows the privacy-accuracy trade-off at $\alpha=0.5$, $\rho=0.1$, and $\sigma_\xi=0.1$, where $E$ and cwOnAveKL are parametrized by $\sigma_\eta$ for (a) \textbf{v1} and (b) \textbf{v2}.
The divergence of cwOnAveKL corresponds to $\sigma_\eta=0$, and moving rightward corresponds to increasing $\sigma_\eta$: privacy initially improves (cwOnAveKL decreases), while excessive noise degrades both privacy and generalization, a behavior similar to that observed in uniform covariate case under isotropic noise \cite{sakata2026privacy}.
Across both settings and a wide range of $\lambda$, the Gram-based perturbation achieves a consistently better trade-off than isotropic perturbation, with lower $E$ and cwOnAveKL and a minimum closer to the origin.

\section{Conclusion and Discussion}

We analyzed private LASSO with objective perturbation under heterogeneous
covariate scales, showing that uniform perturbation can destabilize AMP
and degrade utility, whereas Gram-based perturbation improves stability,
prediction accuracy, and privacy.

While aligning noise with the Gram matrix requires global statistics and may consume additional privacy budget, our results suggest that appropriate perturbation design can yield substantial gains in stability and utility. 
An important direction for future work is to develop data-independent anisotropic designs, for example by leveraging structural priors or public auxiliary data, avoiding additional budget consumption.

For general non-diagonal covariance structures where standard AMP may fail, extending the analysis to Vector AMP (VAMP) \cite{ma2017orthogonal, rangan2019vector} is a natural next step. 
The rotational invariance of VAMP provides a more robust perturbation framework for handling general design matrices.

The idea of injecting noise into the objective is closely related to the Perturb-and-MAP framework \cite{papandreou2011perturb} and MAP-based sampling methods, suggesting a connection to stability analysis under structured noise. 
Moreover, the interplay between noise design and covariate scale is related to compressive sensing with matrix uncertainty \cite{parker2011compressive}, and may enable robust estimation beyond the privacy setting. 
Connections to geometry-aware optimization, such as natural gradient methods, are also worth exploring.

\appendices

\section{Derivation of the stability condition}
\label{sec:app_AT}

The stability of the AMP fixed point can be analyzed via linear stability analysis. 
Let $\widehat{\bm{x}}$ be a fixed point of AMP with the corresponding $\widehat{\bm{\theta}}$ in Algorithm~\ref{alg:AMP}. 
We introduce a small deviation and track its propagation through the AMP updates.
A perturbation in $\widehat{\bm{x}}$ propagates to $\widehat{\bm{\theta}}$ and back as
\begin{align*}
    \delta\widehat{\theta}_\mu^2\simeq\frac{1}{N}\sum_{i=1}^Nv_i\delta\widehat{x}_i^2,~~\delta\widehat{x}_i^2\simeq \frac{\alpha s_{x,i}^2v_i}{(1+s_\theta)^2}\left(\frac{1}{M}\sum_{\mu=1}^M\delta\widehat{\theta}_\mu^2\right).
\end{align*}
Combining the above relations,
$D:=\frac{1}{N}\sum_{i=1}^Nv_i\delta\widehat{x}_i^2$ satisfies
\begin{align}
    D\simeq \frac{\alpha}{(1+s_\theta)^2}\left(\frac{1}{N}\sum_{i=1}^Nv_i^2s_{i}^2\right) D.
\end{align}
Thus, the fixed point is locally stable (i.e., $D \to 0$) when the coefficient is smaller than one. 
Using the expression of $s_{i}$ in Algorithm~\ref{alg:AMP}, the coefficient can be written as
\[
\frac{1}{\alpha N}\sum_{i=1}^N \mathbb{I}(|\mathrm{m}_i - \eta_i \Sigma_i| > \lambda \Sigma_i)\xrightarrow{N\to\infty}\frac{\mathbb{E}_{{\cal D},\bm{\eta}}[\widehat{\rho}]}{\alpha}.
\]
This yields the stability condition $\mathbb{E}_{{\cal D},\bm{\eta}}[\widehat{\rho}] < \alpha$.

\section{Derivation of the State evolution equation}
\label{app:state_evolution}

Following the same line of analysis in \cite{zdeborova2016statistical}, we obtain 
$\mathbb{E}[\mathrm{m}_i]=x_i^{(0)}$
and
\[
    \mathbb{E}_{\xi,F}\!\biggl[\left(\mathrm{m}_i^{(t)}\!-\!x_i^{(0)}\right)^2\biggr]\!=\!\frac{1}{\alpha v_i}\!\biggl\{\frac{1}{N}\sum_{j\neq i}v_j(x_j^{(0)}-\widehat{x}_j^{(t)})^2+\sigma_\xi^2\biggr\}.
\]
In the high-dimensional limit, the empirical average over $j$ concentrates to its expectation with respect to the underlying distribution of $(v, x^{(0)}, \eta)$.
Hence,
\begin{align}
    \mathbb{E}_{\xi,F}[(\mathrm{m}_i^{(t)}-x_i^{(0)})^2]=E^{(t)}\slash(\alpha v_i),
\end{align}
where $E^{(t)}$ corresponds to generalization error at iteration $t$.

\bibliographystyle{IEEEtran}
\bibliography{reference_HeteroPrivateLASSO} 


\end{document}